\documentclass[sigconf]{acmart}
\usepackage{algorithm, algorithmic}
\usepackage{makecell}
\usepackage{multirow}
\usepackage{subfigure}
\usepackage{adjustbox}

\AtBeginDocument{%
  \providecommand\BibTeX{{%
    \normalfont B\kern-0.5em{\scshape i\kern-0.25em b}\kern-0.8em\TeX}}}

\setcopyright{acmlicensed}
\copyrightyear{2024}
\acmYear{2024}
\acmDOI{XXXXXXX.XXXXXXX}

\acmISBN{978-1-4503-XXXX-X/18/06}

\begin{document}

\title{Determined Multi-Label Learning via Similarity-Based Prompt}

\author{Meng Wei}
\email{weimeng@cumt.edu.cn}
\orcid{0009-0000-3836-6487}
\affiliation{%
	\institution{China University of Mining and Technology}
	\city{XuZhou}
	\state{JiangSu}
	\country{China}
}

\author{Zhongnian Li}
\email{zhongnianli@cumt.edu.cn}
\affiliation{%
	\institution{China University of Mining and Technology}
	\city{XuZhou}
	\state{JiangSu}
	\country{China}
}

\author{Peng Ying}
\email{yingpeng@cumt.edu.cn}
\affiliation{%
	\institution{China University of Mining and Technology}
	\city{XuZhou}
	\state{JiangSu}
	\country{China}
}

\author{Yong Zhou}
\email{yzhou@cumt.edu.cn}
\affiliation{%
	\institution{China University of Mining and Technology}
	\city{XuZhou}
	\state{JiangSu}
	\country{China}
}

\author{Xinzheng Xu}
\email{xxzheng@cumt.edu.cn}
\affiliation{%
	\institution{China University of Mining and Technology}
	\city{XuZhou}
	\state{JiangSu}
	\country{China}
}

\renewcommand{\shortauthors}{Meng Wei and Zhongnian Li, et al.}

\begin{abstract}
  In multi-label classification, each training instance is associated with multiple class labels simultaneously. Unfortunately, collecting the fully precise class labels for each training instance is time- and labor-consuming for real-world applications. To alleviate this problem, a novel labeling setting termed \textit{Determined Multi-Label Learning} (DMLL) is proposed, aiming to effectively alleviate the labeling cost inherent in multi-label tasks. In this novel labeling setting, each training instance is associated with a \textit{determined label} (either "Yes" or "No"), which indicates whether the training instance contains the provided class label. The provided class label is randomly and uniformly selected from the whole candidate labels set. Besides, each training instance only need to be determined once, which significantly reduce the annotation cost of the labeling task for multi-label datasets. In this paper,  we theoretically derive an risk-consistent estimator to learn a multi-label classifier from these determined-labeled training data. Additionally, we introduce a similarity-based prompt learning method for the first time, which minimizes the risk-consistent loss of large-scale pre-trained models to learn a supplemental prompt with richer semantic information. Extensive experimental validation underscores the efficacy of our approach, demonstrating superior performance compared to existing state-of-the-art methods.
\end{abstract}




\received{20 February 2007}
\received[revised]{12 March 2009}
\received[accepted]{5 June 2009}

\maketitle

\section{Introduction}
 In multi-label  learning (MLL) tasks, each training instance is equipped with multiple class labels simultaneously  \cite{ML_1, ML_2, ML_3, ML_4, ML_5, ML_6}. For example, a new post shared on Instagram often has multiple hashtags; a user-uploaded image often contains multiple individuals with different characteristics \cite{ML_7, ML_8, ML_9}. However, in practice, the initial data we collect usually has rich semantics, diverse scenes, and multiple targets \cite{SPML_1, SPML_2, SPML_3}. If all the targets in this data are to be accurately labeled, it undoubtedly requires significant human resources, especially when the potential label space is very large, as labeling these training data typically demands both manpower and expertise \cite{SPML_4, SPML_5}. 

 So how to investigate an alternative approach to alleviate this problem, freeing annotators from the heavy task of labeling, is a crucial issue \cite{SPML_1, SPML_2, ML_7}. During the past decade, numerous studies have made great efforts to alleviate this issue, demonstrating that even weaker annotations can yield satisfactory results in multi-label tasks. These studies including but not limited to multi-label missing labels (MLML) \cite{ML_2, ML_13, ML_14}, partial multi-label learning (PML) \cite{PLML_1, PLML_2}, and complementary multi-label learning (CML) \cite{CMLL}. In the latest research, the single positive multi-label learning assumes that each training instance is equipped with only one positive labels. 
 
  \begin{figure*}[!htbp]
 	\centering
 	\includegraphics[width=\linewidth]{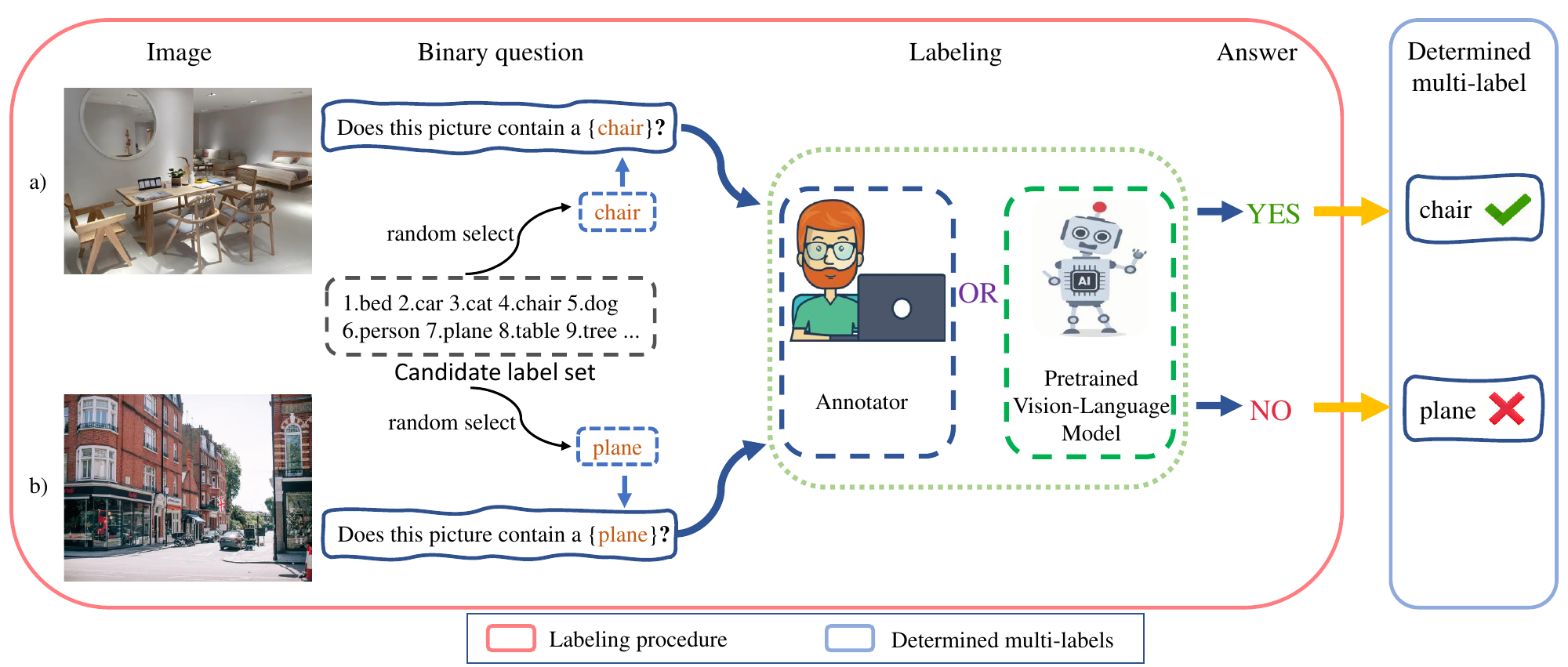}
 	\caption{An example of determined multi-label learning. a) A image with "chair" label; b) A image without "plane" label. Compared to precisely annotating all the relevant labels, determining the presence of a randomly generated label in the image is undoubtedly easier and more time-efficient. This labeling procedure requires only a single assessment per image, making it suitable for labeling tasks on the future real-world large-scale datasets.}
 	\label{figure_1}
 \end{figure*}
  
 In this paper, we explore a novel labeling setting termed as \textit{determined multi-label}, which significantly reduces the labeling cost for multi-label datasets. In this novel setting, the \textit{determined multi-label} indicates whether the training instance contains a randomly provided class label. Specifically, the \textit{determined multi-label} includes two case: either "Yes" (i.e., the training instance contains the provided class label), or "No" (i.e., the class label is irrelevant). It's evident that answering a "Yes or No" question is simpler than answering a multiple choices question. Compared to precisely labeling all the relevant class labels from an extensive candidate classes set, determining one label is simple and significantly reduces the labeling cost for multi-label tasks.  For example, as illustrated in Figure \ref{figure_1}, for a room image, annotators can easily determine that the room image contains the "chair" label. Similarly, for a street image, annotators can quickly recognize that the "plane" label is irrelevant. Therefore, \textit{determined multi-label} can significant reduce the time cost for browsing the candidate labels set. To sum up, \textit{determined multi-label} is a worthwhile problem for at least these two reasons: Firstly, this novel labeling setting can significant reduce the labeling cost for multi-label tasks. Secondly, this setting is friendly to large-scale vision-language models, because it is particularly suitable for designing simple prompts for labeling, as illustrated in Figure \ref{figure_1} .

 We first propose the determined multi-label Learning (DMLL) method through similarity-based prompt. Specifically, first, we propose a theoretically guaranteed risk-consistent loss, which leverages large-scale pre-trained models to extract image and text features, generate estimated probabilities for $p(y^j=1\mid x)$, and rewrite multi-label classification risk based on determined-labeled data. Second, we introduce a similarity-based prompt learning method, which minimizes the risk-consistent loss to learn a supplemental prompt with richer semantic information. We make three main contributions: 

 1. A novel labeling setting for multi-label datasets is introduced, which effectively alleviate the annotation  cost of multi-label tasks.

 2. A risk-consistent loss tailed for this new setting is proposed with theoretical guarantees. 

 3. A novel similarity-based prompt learning method is introduced for the first time, which enhances the semantic information of class labels by minimize the proposed risk-consistent loss. 

 Excitingly, experimental results demonstrate that our method outperforms the existing state-of-the-art weakly multi-label learning methods significantly. We believe that this exploration of large-scale model for weak supervision has the potential to promote further development in the weak supervision domain, making effective use of extremely limited supervision possible.

\section{Related Work}
 In this paper, we present the relevant references, which can be categorized into two main areas: vision-language model and weakly multi-label learning.
\subsection{Vision-Language Model}
 Large-scale Vision-Language Model (VLM) have achieved significant success in various downstream visual tasks, thanks to their outstanding zero-shot transfer capabilities and rich prior knowledge \cite{VLM1, VLM2, VLM3, VLM4, VLM5}. Among them, the most commonly used pre-trained VLM is Contrastive Language-Image Pre-Training (CLIP) \cite{CLIP}. Recently, researchers have attempted to enhance the generalization capability of models by leveraging the prior knowledge embedded in large-scale VLM \cite{VLM6, VLM7, VLM8, VLM9, VLM10}. Dual-CoOp \cite{DUALCoOp} learns the relationships between different category names by aligning image and CLIP text space in multi-label tasks. They learn a positive and negative "prompt context" for potential targets, then compute similarity based on local features of the image to achieve multi-label recognition. SCPNet \cite{SCPnet} utilizes the CLIP model to extract an object association matrix as prior information to better understand the connections between multiple labels in an image. MKT \cite{MKT} uses CLIP's initial weights for knowledge transfer to enhance zero-shot multi-label learning. VLPL \cite{VLPL}, based on CLIP image-text similarity, extracts positive and pseudo-labels, and ultilizes entropy maximization loss to improve performance under the Single Positive Multi-Label Learning (SPML) setting. 

  \begin{figure*}[!htbp]
	\centering
	\includegraphics[width=\linewidth]{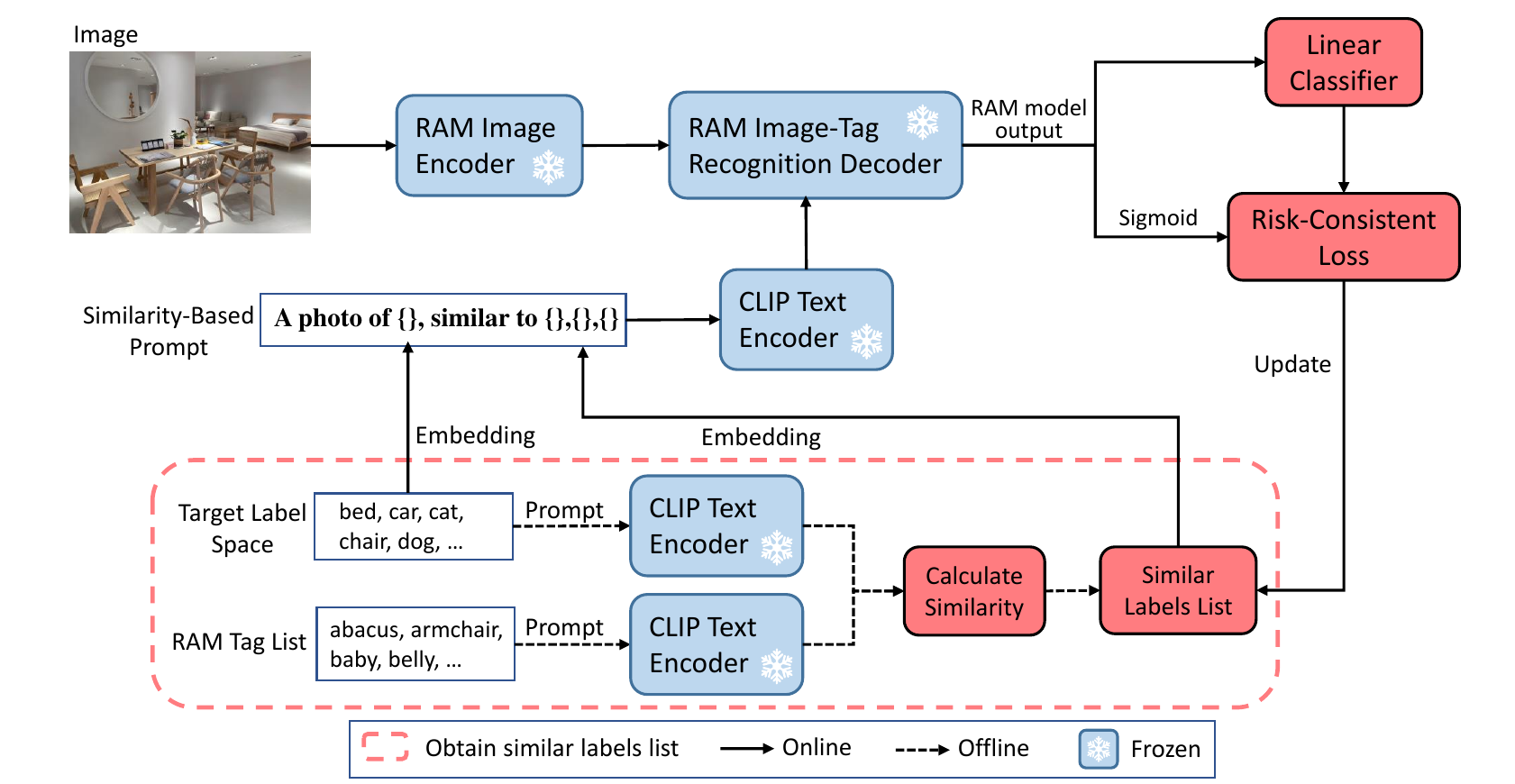}
	\caption{The architecture of the proposed model. We design a similarity-based prompt (SBP) strategy to enhance the inherent semantic information of labels. First, we offline generate a similar labels list for the target labels set based on the large labels set organized by RAM. Second,  these similar labels are embedded into a the proposed SBP. Subsequently, we utilize the RAM image encoder and CLIP text encoder to extract image and text features. Finally, we optimize the proposed risk-consistent loss based on the RAM model output and the linear classifier output, and iteratively update the similar labels list.}
	\label{figure_2}
\end{figure*}

 However, these CLIP-based methods show unsatisfactory performance on multi-label tasks, as it is trained on large-scale multi-class datasets (i.e., each image is associated with one label). Therefore, this paper employs the Recognize Anything Model  (RAM) \cite{RAM} for image feature extraction. RAM is train on large-scale multi-label datasets, demonstrating advanced zero-shot transfer capabilities on predefined categories. Therefore, using the image encoder of RAM can extract richer semantic information for multi-label tasks. Additionally, we utilize the large-scale image labels set of RAM to learn a similarity-based supplemental prompt, further enhancing the semantic information of labels and achieving performance improvement in multi-label tasks.
\subsection{Weakly Multi-Label Learning}
 Significant progress has been made in multi-label learning in recent years, thanks to the collection of fully annotated multi-label datasets \cite{ML_8, ML_9, ML_10, ML_11, ML_12, ML_14, ML_15}. However, as the scale of label sets expands, the task of collecting fully annotations for large-scale datasets becomes increasingly challenging and labor-intensive. To alleviate this problem, researchers actively investigate how to reduce the annotation cost in multi-label tasks, i.e., weakly supervised multi-label scenarios. Partial multi-label learning (PML), as an alternative to fully annotation, provides a label set for each image containing some correct labels (with some noise) \cite{PLML_1, PLML_2, ML_15}. Xie et al. disambiguate between ground-truth and noisy labels in a meta-learning fashion \cite{PLML_1}. Itamar et al. propose to estimate the class distribution using a dedicated temporary model \cite{PLML_2}. These methods alleviate the need for expert knowledge in annotation, allowing annotators to make errors.
 
 Recently, a highly attractive setting--Single Positive Multi-Label Learning (SPML) \cite{SPML_1}, assumes each image is equipped with only one positive label. For SPML tasks, much of the work focuses on developing novel loss functions to train models. Assume Negative (AN) Loss \cite{SPML_1}, widely compared as a baseline, assumes that all unknown labels are negative, but it inevitably introduces some false negatives during implementation. Entropy-Maximization (EM) Loss \cite{SPML_2} maximizes the entropy of predicted probabilities for unknown labels. Weak Assume Negative (WAN) Loss \cite{SPML_1} is an improved version of the AN Loss, providing a relatively weak coefficient weight for negative labels, thereby reducing the impact of false negatives. Regularized Online Label Estimation (Role) Loss \cite{SPML_1} estimates the labels of unknown labels online and utilize binary entropy loss to align classifier predictions with estimated labels. Large Loss \cite{SPML_3} first learns the representation of clean labels and then memorizes labels with noise.
 
 However, equipping each image with a positive label in real-world scenarios remains unrealistic, as it can be viewed as a weakened version of the multi-class problem. In practice, large-scale multi-class datasets still require a large number of manual annotation cost. Consequently, we consider another novel annotation mechanism to liberate annotators from the arduous annotation task, known as the Determined Multi-Label Learning (DMLL). Specifically, for each image, annotators only need to determine whether  the image contains the provided class label. This type of annotation data is easily obtainable in real life.
\section{The Proposed Method}
 In this section, we provide a detailed description of the proposed Determined Multi-label Learning (DMLL) method. Firstly, we describe the problem definition and discuss the generation of determined multi-label.  Secondly, we describe the proposed risk-consistent loss. Subsequently, we present how the proposed similarity-based supplemental prompt effectively enriches the semantic information of labels, enhancing the model's performance. Finally, we illustrate the architecture of the proposed method in detail.
\subsection{Determined Multi-Label}
Compared to precisely labeling all the relevant labels from a exhaustive set of candidate labels, we consider another scenario--Determined Multi-Label Learning (DMLL), which effectively alleviate the annotation  cost of multi-label tasks. In this novel setting, each training instance is equipped with a randomly assigned binary determined multi-label. Specifically, a class label is randomly and uniformly generated from the candidate label set, and the annotator only needs to determine whether the image contains the provided class label. For example, assuming our candidate labels are $\{$"bed", "building", "bicycle", "car", "cat",  "chair", "dog", "glass", "flower", "person", "plane", "table", "train", "tree", ...$\}$, for an "room" image, if the randomly generated label is "chair" label, the annotator can easily determine that the "chair" label exists in the image, i.e., the label "chair" is relevant to the image. For an "street" image, if the randomly generated label is "plane" label, the annotator can also  easily determine that the "plae" label does not exist, i.e., the label "plane" is irrelevant. In practice, answering a "Yes-or-No" question (i.e., Determined Multi-Label) are simpler than answering a choice question (i.e., Single Positive Multi-Label), not to mention a multi-choice question (i.e., Multi-Label). As expected, this setting significantly reduces annotation costs of labeling multi-label tasks compared to the traditional multi-label learning settings, making it highly suitable for crowdsourced annotation tasks and the labeling tasks for future large-scale datasets.
\subsection{Problem Setup}
\subsubsection{Multi-Label Learning}
 In multi-label learning (MLL) problems, each instance is associated with multiple class labels simultaneously, and aims to build a model that can predict multiple relevant class labels for the unseen examples. Let $ \mathcal{X} \subset \mathbb{R}^{d} $ be the instance space with $d$-dimensions and $\mathcal{Y} = \{1, 2, \ldots, k\}$ be the label space with $k$ class labels. Suppose $\mathcal{D} = \{(x_i, Y_i)\}_{i=1}^{N}$ is the MLL training set where $N$ denotes the number of training samples, $x_i \in \mathcal{X}$ denotes the $d$-dimensional training instance and $Y_i \in \mathcal{C}$ represents the set of multiple relevant class labels of $x_i$ where $\mathcal{C} = 2^{\mathcal{Y}}$. The aim of learning from multi-labels is to train a multi-label classifier $f(x): x \mapsto 2^{\mathcal{Y}}$ that minimizes the following expected risk:
 \begin{equation}\label{eq1}
 	R(f) = \mathbb{E}_{p(x, Y)}[\mathcal{L}(f(x), Y)],
 \end{equation}
 where $\mathbb{E}$ represents the expectation and $\mathcal{L}$ denotes the multi-label loss function.
\subsubsection{Determined Multi-label Learning}
 In this paper, we consider the scenario where each instance is associated with determined label instead of fully annotation. Similarly, let $ \mathcal{X} \subset \mathbb{R}^{d} $ be the $d$-dimensional instance space and $\mathcal{Y} = \{1, 2, \ldots, k\}$ be the label space with $k$ class labels. Suppose the determined-labeled training dataset $\bar{D} = \{(x_i, y_{i}^\gamma)\}_{i=1}^{N}$ is sampled randomly and uniformly from an unknown probability distribution with density $p(x, y^{\gamma})$ where $N$ denotes the number of training samples and $\gamma \in \mathcal{Y}$ denotes the observed determined label of $x_i$. For each determined-labeled sample $(x_i, y_i^{\gamma})$, $y_{i}^{\gamma} \in \{0, 1\}$ denotes the value of determined label $\gamma$. Specifically, $y_{i}^{\gamma} = 1$ demotes the label $\gamma$ is relevant to the instance $ x_i $ and $y_i^{\gamma} = 0 $ means the label $\gamma$ is irrelevant. Similarly, our goal is to learn a multi-label classifier $f(x): x \mapsto 2^{\mathcal{Y}}$ from these determined-labeled samples, which can predict multiple relevant labels for the unseen images.
 
 The determined multi-label learning problem may be solved by the method of single positive multi-label learning, where each instance is associated with one relevant class label---single positive multi-label learning problem can be regarded as a special case of determined multi-label learning where all the $y^{\gamma}$ are equal to $1$, which is not realistic in the real-world datasets.

\subsection{Risk-Consistent Estimator} 
 In this section, based on our proposed problem setup, we present a novel risk-consistent method. To deal with determined multi-label learning problem, the classification risk $R(f)$ in Equation (\ref{eq1}) could be rewritten as 
 \begin{equation}\label{eq2}
 	\begin{aligned}
 		& \mathbb{E}_{p(x, Y)}[\mathcal{L}(f(x), Y)] \\
 		& = \int_{x} \sum_{Y \in \mathcal{C}} p(Y \mid x) \mathcal{L} (f(x), Y) p(x)dx \\
 		& = \int_x \sum_{Y \in \mathcal{C}} \sum_{\gamma \in \mathcal{Y}} \frac{1}{k}\mathcal{L}(f(x), Y)p(Y\mid x)p(x)dx \\
 		& = \int_x \sum_{Y \in \mathcal{C}} \sum_{\gamma \in Y} \frac{1}{k} \mathcal{L}(f(x), Y) \frac{p(Y\mid x)}{p(y^{\gamma}=1\mid x)}p(y^{\gamma}=1\mid x) p(x)dx \\
 		& \qquad + \int_x \sum_{Y \in \mathcal{C}} \sum_{\gamma \notin Y} \frac{1}{k} \mathcal{L}(f(x), Y) \frac{p(Y\mid x)}{p(y^{\gamma}=0\mid x)}p(y^{\gamma}=0\mid x) p(x)dx \\
 		& = \mathbb{E}_{p(x, y^{\gamma} = 1) } \frac{1}{p(y^{\gamma} = 1 \mid x) k} \sum_{Y \in \mathcal{C}} \mathcal{L}(f(x), Y)p(Y\mid x) \\
 		& \qquad + \mathbb{E}_{p(x, y^{\gamma} = 0) } \frac{1}{p(y^{\gamma} = 0 \mid x) k} \sum_{Y \in \mathcal{C}} \mathcal{L}(f(x), Y)p(Y\mid x) \\
 		& = R_{d}(f),
 	\end{aligned}
 \end{equation}
 where $R_d(f)$ denotes the classification risk of determined multi-label learning.
 
  \begin{algorithm}[htbp]
 	\caption{Determined Multi-Label Learning}
 	\label{alg:Framwork}
 	\begin{algorithmic}[1]
 		\REQUIRE ~~\\ 
 		The determined labeled training set $\bar{D} = \{(x_i, y_{i}^\gamma)\}_{i=1}^{N}$, where $ y_{i}^\gamma = 1 $ and $y_{i}^\gamma=0 $ denotes whether the label $ \gamma $ belongs to the instance $x_i$ or not; \\
 		The number of epochs, $ T $; \\
 		The optimal similarity-based prompt, $P^*$;\\
 		The number of epochs for update $P^*$, $M$;\\
 		An external stochastic optimization algorithm, $ \mathcal{A} $;
 		\ENSURE ~~\\ 
 		model parameter $ \theta $ for $f(x, \theta)$;
 		\STATE \textbf{Initialize} $P^*$ with zero similar label according to Eq.(\ref{eq9});
 		\FOR{$t=1$ to $T$}
 		\STATE \textbf{Shuffle} $\bar{D} = \{(x_i, y_{i}^\gamma)\}_{i=1}^{N}$ into $B$ mini-batches;
 		\FOR{$ b=1 $ to $ B $}
 		\STATE Fetch mini-batch $\bar{D}_b$ from $\bar{D}$;
 		\STATE Fixed $P^*$, update model parameter $ \theta $ by $\hat{R}_d$ in Eq.(\ref{eq7});
 		\ENDFOR
 		\IF{t $\%$ M == 0} 
 		\STATE Fixed $f(x,\theta)$, update $P^*$ according Eq.(\ref{eq10})
 		\ENDIF
 		\ENDFOR
 	\end{algorithmic}
 \end{algorithm}
 
 Subsequently, for the multi-label loss function $\mathcal{L}$, we utilize the widely used binary cross-entropy loss function in multi-label learning, thus we can obtain the following formula:
 \begin{equation}\label{eq3}
 	\begin{aligned}
 		\mathcal{L}(f(x), Y) = \sum_{j \in Y} log (f_j(x)) + \sum_{j \notin Y}log(1-f_j(x)).
 	\end{aligned}
 \end{equation}
 For ease of reading, let $\ell^{j} = log(f_j(x))$ and $\bar{\ell}^j = log(1-f(x))$, we can obtain:
 \begin{equation}\label{eq4}
 	\mathcal{L}(f(x), Y) = \sum_{j\in Y} \ell^j + \sum_{j \notin Y} \bar{\ell}^j.
 \end{equation}
 Accordingly, 
  $\sum_{Y \in \mathcal{C}} \mathcal{L}(f(x), Y)p(Y\mid x)$ in Equation  (\ref{eq2}) can be calculated by:
 \begin{equation}\label{eq5}
 	\begin{aligned}
 		& \sum_{Y \in \mathcal{C}} \mathcal{L}(f(x), Y)p(Y\mid x) \\
 		& \qquad \qquad \qquad = \sum_{j=1}^{k}[p(y^j = 1 \mid x) \ell^j + (1-p(y^j \mid x))\bar{\ell}^j] \\
 		& \qquad \qquad \qquad = H_j .
 	\end{aligned}
 \end{equation}
 Here, $p(y^j = 1\mid x)$ could be regarded as the soft label for class $j$ to $x$. The proof is given in Appendix.  Consequently, by substituting Equation (\ref{eq5}) into Equation (\ref{eq2}), a risk-consistent estimator for DMLL can be derived by the following formula:
 \begin{equation}\label{eq6}
 	\begin{aligned}
 		R_d(f) & = \mathbb{E}_{p(x, y^{\gamma} = 1) } \left[\frac{1}{p(y^{\gamma} = 1 \mid x) k} H_j \right] \\
 		& \qquad\qquad + \mathbb{E}_{p(x, y^{\gamma} = 0) } \left[\frac{1}{p(y^{\gamma} = 0 \mid x) k} H_j \right].
 	\end{aligned}
 \end{equation}
 Since the training dataset $\bar{D} = \{(x_i, y_{i}^\gamma)\}_{i=1}^{N}$ is sampled randomly and uniformly from the $p(x, y^{\gamma})$, the empirical risk estimator can be expressed as:
 \begin{equation}\label{eq7}
 	\begin{aligned}
 		\hat{R}_d(f) & = \frac{1}{N_{y^{\gamma} = 1}} \sum_{i = 1}^{N_{y^{\gamma} = 1}} \left(\frac{1}{p(y_i^{\gamma} = 1 \mid x_i)k}H_j\right) \\
 		& \qquad\qquad + \frac{1}{N_{y^{\gamma} = 0}} \sum_{i = 1}^{N_{y^{\gamma} = 0}} \left(\frac{1}{p(y_i^{\gamma} = 0 \mid x_i)k}H_j\right),
 	\end{aligned}
 \end{equation}
 where $N_{y^{\gamma} = 1}$ and $N_{y^{\gamma} = 0}$ denote the number of training sample with determined-label $y^{\gamma} = 1$ and $y^{\gamma} = 0$. Actually, since the probability of $d^j = p(y^j = 1 \mid x)$ is not available from the given data, we apply the sigmoid function on the RAM model output $f(x)$ to recover the soft label $d^j$ and $\bar{d}^j$, which can also be found in other weakly supervised setting \cite{PLL1, SPML_4}. Besides, we utilize the limited supervision generated by the determined label to correct the output of the RAM model and thus better estimate the probability of $p(y^j = 1 \mid x)$. 
\subsection{Similarity-Based Prompt}
 Recently, large-scale cross-modality pretrained models have achieved remarkable success in visual tasks. Researchers explore various prompt learning methods to utilize large-scale models efficiently for downstream visual tasks, e.g., CoCoOp \cite{CoCoOp}, HSPNet \cite{HSPNet}. In this paper, we propose a simple yet efficient similarity-based supplemental prompt learning method to enhance the semantic information of labels. We actively select labels similar to those of the target labels to expand the prompt from the big labels set space provided by RAM. The supplemental prompt not only enriches the semantic information of labels but also enhances the overall semantic information of sentences.
 
 Specifically, given a target label space $ Y_T =\{y_1, y_2, \ldots, y_k\}$ and a large label space $Z_L=\{z_1, z_2, \ldots, z_l\}$ where $k$ and $l$ denotes the number of $Y_T$ and $Z_L$. Here, we utilize the labels set provided in the RAM paper for $Z_L$, which contains 4585 classes, including some rare labels. First, We introduce a fixed prompt template, i.e., "a photo of a $\{$class$\}$", where the "$\{$class$\}$" represents the name of the given label. We utilize the text encoder of the CLIP model, denoted as E(·) to extract text features. For ease of explanation, let $u_k$ and $v_l$ be the word embedding of the prefix "a photo of a $\{$class$\}$" from the $Y_T$ and $Z_L$. The similarity between $u_k$ and $v_l$ can be calculated by:
 \begin{equation}\label{eq8}
 	S(k, l) = sim\left[E(u_k), E(v_l)\right],
 \end{equation}
 where $S(k, l)$ denotes the similarity between the label $y_k$ in $Y_T$ and the label $z_l$ in $Z_L$. 
 
 Second, we actively select top $\sigma$ similar labels from $Z_L$ for each label in $Y_T$. Second, we introduce a similarity-based prompt, "a photo of a $\{y_k\}$, similar to $\{z_1, z_2, \ldots, z_{\lambda}\}$", to enhance the semantic information of labels $y_k$, where $\lambda \leq \sigma$. Let $P = \{P_1^{\lambda}, P_2^{\lambda}, \ldots, P_k^{\lambda}\}$ denotes the similarity-based prompt for each label in $Y_T$. The similarity-based prompt $P_j^{\lambda}$ can be expressed as:
 \begin{equation}\label{eq9}
 	P_j^{\lambda}= \left\{
 	\begin{aligned}
 		G_j, \qquad \qquad \qquad \lambda=0 \\
 		G_j \left[w_0 \divideontimes w_1\divideontimes \cdots \divideontimes w_{\lambda}\right], \qquad \lambda \neq 0 \\
 	\end{aligned}
 	\right
 	.,
 \end{equation}
 where $G_j$ is the word embedding of the prefix "a photo of a $\{y_j\}$", $w_0$ denotes "similar to", and $w_{\lambda}$ denotes the $\lambda$-th similar label for $y_j$. Here, $\divideontimes$ connects multiple similar labels for $y_j$. 
 
   \begin{table}[!htbp]
 	\centering
 	\renewcommand{\arraystretch}{1}
 	\caption{The details of four benchmark datasets.}
 	\begin{tabular}{l|c|c|c|c}%c表示文本居中，c的个数是列数
 		\toprule[1pt]
 		Dataset & $\#$Classes & $\#$training set  & \makecell{$\#$Positive\\($y^{\gamma}=1$))} & \makecell{$\#$Negative\\($y^{\gamma}=0$)}  \\
 		\midrule
 		VOC & 20 & 5717 & 6.9$\%$ & 93.1$\%$ \\
 		\midrule
 		COCO & 80 & 82081 & 3.6$\%$ & 96.4$\%$ \\
 		\midrule
 		NUS &  81 & 161789 & 2.3$\%$ & 97.7$\%$ \\
 		\midrule
 		CUB & 312 & 5994 & 9.8$\%$ & 90.2$\%$ \\
 		\bottomrule[1pt]
 	\end{tabular}
 	\label{table_1}
 \end{table}
 
 Subsequently, we minimize the risk-consistent loss, as proposed in the previous section, from a batch of data $B=\{(x_i,y_i^{\gamma})\}_{i=1}^{b}$, to obtain the optimal $P^*$. This optimization process involves iterating over each $P_j^{\lambda}$. Through this iteration, the optimal $P^*$ can be expressed as follows:
 \begin{equation}\label{eq10}
 	P^* = \mathop{argmin}_{\lambda} \sum_{i=1}^{b}\mathcal{L}(f(x, P), Y).
 \end{equation}
 Here, we minimize the risk-consistent loss to learn the optimal $\lambda$ similar labels for each label from $Y_T$.
 
 \begin{table*}[!htbp]
 	\centering
 	\renewcommand{\arraystretch}{1.2}
 	\tabcolsep=0.1em
 	\caption{Comparison results between the proposed method and state-of-the-art weakly multi-label learning methods (mean$\pm$std) in terms of \textit{MAP} (the greater, the better) and \textit{One error} (the smaller, the better). The best performance is in bold. }
 	\begin{tabular*}{\textwidth}{@{\extracolsep{\fill}}l|cccc|cccc}%c表示文本居中，c的个数是列数
 		\toprule[1pt] 
 		\multirow{2}{*}{\textbf{Method}} & \multicolumn{4}{c|}{MAP $\uparrow$} & \multicolumn{4}{c}{One error $\downarrow$} \\
 		\cline{2-9}
 		~ & VOC & COCO & NUS & CUB & VOC & COCO & NUS & CUB\\
 		\midrule
 		AN \cite{SPML_1} & 0.836$\pm$0.005 & 0.536$\pm$0.007 & 0.234$\pm$0.001 & 0.145$\pm$0.001 & 0.118$\pm$0.006 & 0.245$\pm$0.012 & 0.532$\pm$0.001 & 0.448$\pm$0.002 \\
 		WAN \cite{SPML_1} & 0.825$\pm$0.004 & 0.528$\pm$0.001 & 0.231$\pm$0.001 & 0.141$\pm$0.001 & 0.120$\pm$0.002 & 0.223$\pm$0.001 & 0.535$\pm$0.002 & 0.481$\pm$0.015 \\
 		ROLE \cite{SPML_1} & 0.847$\pm$0.004 & 0.580$\pm$0.000 & 0.241$\pm$0.002 & 0.126$\pm$0.001 & 0.077$\pm$0.001 & 0.202$\pm$0.017 & 0.535$\pm$0.001 & 0.418$\pm$0.001 \\
 		EM \cite{SPML_2} & 0.842$\pm$0.001 & 0.586$\pm$0.003 & 0.270$\pm$0.016 & 0.138$\pm$0.002 & 0.077$\pm$0.001 & 0.180$\pm$0.003 & 0.513$\pm$0.001 & 0.437$\pm$0.015 \\
 		EM$\_$APL \cite{SPML_2} & 0.855$\pm$0.012 & 0.592$\pm$0.002 & 0.275$\pm$0.003 & 0.143$\pm$0.001 & 0.067$\pm$0.001 & 0.169$\pm$0.010 & 0.425$\pm$0.003 & 0.420$\pm$0.001 \\
 		Scob \cite{SPML_5} & 0.841$\pm$0.011 & 0.594$\pm$0.001 & 0.362$\pm$0.008 & 0.143$\pm$0.002 & 0.071$\pm$0.001 & 0.154$\pm$0.001 & 0.567$\pm$0.002 & 0.438$\pm$0.009 \\
 		VLPL \cite{VLPL} & 0.871$\pm$0.003 & 0.688$\pm$0.001 & 0.411$\pm$0.008 & 0.169$\pm$0.001 & 0.075$\pm$0.003 & 0.054$\pm$0.001 & 0.426$\pm$0.007 & 0.480$\pm$0.024 \\
 		PLCSL \cite{PLML_2} & 0.857$\pm$0.002 & 0.615$\pm$0.002 & 0.278$\pm$0.001 & 0.144$\pm$0.001 & 0.092$\pm$0.003 & 0.155$\pm$0.009 & 0.429$\pm$0.003 & 0.504$\pm$0.015 \\
 		CLML \cite{CMLL}  &0.680$\pm$0.003 & 0.151$\pm$0.004 & 0.121$\pm$0.007 & 0.121$\pm$0.001 & 0.140$\pm$0.005 & 0.581$\pm$0.028 & 0.547$\pm$0.024 & 0.713$\pm$0.007 \\
 		\midrule
 		DMLL (Ours) & \textbf{0.921$\pm$0.001} & \textbf{0.746$\pm$0.000} & \textbf{0.462$\pm$ 0.003} & \textbf{0.228$\pm$0.001} & \textbf{0.046$\pm$0.001} & \textbf{0.040$\pm$0.001} & \textbf{0.418$\pm$0.001} & \textbf{0.386$\pm$0.003} \\
 		\bottomrule[1pt]
 	\end{tabular*}
 	\label{table_2}
 \end{table*}
 
 \begin{table*}[!htbp]
 	\centering
 	\renewcommand{\arraystretch}{1.2}
 	\tabcolsep=0.1em
 	\caption{Comparison results between the proposed method and state-of-the-art weakly multi-label learning methods (mean$\pm$std) in terms of \textit{Ranking loss} (the smaller, the better) and \textit{Coverage} (the smaller, the better). The best performance is in bold. }
 	\begin{tabular*}{\textwidth}{@{\extracolsep{\fill}}l|cccc|cccc}%c表示文本居中，c的个数是列数
 		\toprule[1pt] 
 		\multirow{2}{*}{\textbf{Method}} & \multicolumn{4}{c|}{Ranking loss $\downarrow$} & \multicolumn{4}{c}{Coverage $\downarrow$} \\
 		\cline{2-9}
 		~ & VOC & COCO & NUS & CUB & VOC & COCO & NUS & CUB\\
 		\midrule
 		AN \cite{SPML_1} & 0.025$\pm$0.001 & 0.044$\pm$0.001 & 0.078$\pm$0.002 & 0.447$\pm$0.002  & 0.058$\pm$0.001 & 0.218$\pm$0.001 & 0.091$\pm$0.001 & 0.855$\pm$0.002 \\
 		WAN \cite{SPML_1} & 0.027$\pm$0.001 & 0.041$\pm$0.003 & 0.073$\pm$0.001 & 0.481$\pm$0.015 & 0.062$\pm$0.002 & 0.214$\pm$0.002 & 0.092$\pm$0.001 & 0.833$\pm$0.003 \\
 		ROLE \cite{SPML_1} & 0.022$\pm$0.000 & 0.038$\pm$0.003 & 0.072$\pm$0.003 & 0.419$\pm$0.001 & 0.057$\pm$0.001 & 0.211$\pm$0.002 & 0.091$\pm$0.000 & 0.876$\pm$0.006 \\
 		EM \cite{SPML_2} & 0.024$\pm$0.001 & 0.037$\pm$0.002 & 0.075$\pm$0.003 & 0.437$\pm$0.015 & 0.061$\pm$0.001 & 0.210$\pm$0.001 & 0.100$\pm$0.003 & 0.950$\pm$0.005 \\
 		EM$\_$APL \cite{SPML_2} & 0.023$\pm$0.001 & 0.036$\pm$0.001 & 0.074$\pm$0.001 & 0.420$\pm$0.001 & 0.059$\pm$0.002 & 0.208$\pm$0.002 & 0.093$\pm$0.001 & 0.841$\pm$0.001 \\
 		Scob \cite{SPML_5} & 0.025$\pm$0.000 & 0.035$\pm$0.001 & 0.098$\pm$0.007 & 0.438$\pm$0.009 & 0.063$\pm$0.000 & 0.207$\pm$0.001 & 0.152$\pm$0.011 & 0.846$\pm$0.006 \\
 		VLPL \cite{VLPL} & 0.029$\pm$0.001 & 0.064$\pm$0.001 & 0.084$\pm$0.013 & 0.680$\pm$0.024 & 0.070$\pm$0.002 & 0.201$\pm$0.001 & 0.139$\pm$0.023 & 0.837$\pm$0.015 \\
 		PLCSL \cite{PLML_2} & 0.021$\pm$0.001 & 0.034$\pm$0.001 & 0.076$\pm$0.001 & 0.503$\pm$0.014 & 0.056$\pm$0.002 & 0.273$\pm$0.001 & 0.094$\pm$0.001 & 0.884$\pm$0.002 \\
 		CLML \cite{CMLL} & 0.051$\pm$0.000 & 0.347$\pm$0.002 & 0.097$\pm$0.011 & 0.375$\pm$0.002 & 0.100$\pm$0.001 & 0.567$\pm$0.004 & 0.143$\pm$0.014 & 0.934$\pm$0.001 \\
 		\midrule
 		DMLL (Ours) & \textbf{0.011$\pm$0.000} & \textbf{0.033$\pm$0.001} & \textbf{0.048$\pm$0.002} & \textbf{0.339$\pm$0.002} & \textbf{0.041$\pm$0.001} & \textbf{0.174$\pm$0.001} & \textbf{0.064$\pm$0.002} & \textbf{0.816$\pm$0.003} \\
 		\bottomrule[1pt]
 	\end{tabular*}
 	\label{table_3}
 \end{table*}
 
 Finally, we fix the optimal similarity-based prompt $P^*$ and optimize the multi-label classifier $f(x)$. To provide a comprehensive understanding of the proposed method, Algorithm 1 illustrates the overall algorithmic procedure, and Figure 2 illustrates the training process.
 
\section{Experiments}
 In this section, we conduct extensive experiments to verify the performance of the proposed method and compare its effectiveness to the existing state-of-the-art weakly multi-label learning methods. For ease of reading, our approach is defined as DMLL.
\subsection{Experimental Setup}
\subsubsection{Datasets} 
 For determined multi-label learning (DMLL), we conduct experiments on four standard benchmark datasets, including PASCAL VOC 2012 (VOC) \cite{VOC}, MS-COCO 2014 (COCO) \cite{COCO}, NUS-WIDE (NUS) \cite{NUS}, and CUB-200-2011 (CUB) \cite{CUB}. To generate determined multi-label for each image in the training datasets, we randomly select a candidate label $\gamma$ from the candidate labels set $\mathcal{Y}$. Subsequently, we validate this label $\gamma$ against the positive labels set $Y$ of the image. Specifically, if $\gamma$ belongs to $Y$, then $y^{\gamma}$ is set to $1$; otherwise, $y^{\gamma}$ is set to $0$. In this setup, the number of positive samples generated in the final training set is significantly smaller than the number of negative samples. This precisely indicates that the proposed setup represents an extreme form of weakly multi-label learning. The more details of these datasets can be obtained from Table \ref{table_1}.
\subsubsection{Compared methods}

 To assess the efficacy of the proposed approach, a thorough evaluation is conducted through comparisons with state-of-the-art weakly multi-label learning methods, including single positive multi-label learning (SPML) methods, partial multi-label learning (PML) methods, and complementary multi-label learning (CML) methods. The key summary statistics for the compared methods are as follows:
 \begin{enumerate}
 	\item [$ \bullet $] AN \cite{SPML_1}: A SPML method assuming all unknown labels are negative.
 	\item [$ \bullet $] WAN \cite{SPML_1}: An enhanced version of AN, featuring a relatively weak coefficient weight assigned to negative labels.
 	\item [$ \bullet $] ROLE \cite{SPML_1}: A SPML approach estimating unknown labels online and leveraging binary entropy loss to align classifier predictions with estimated labels.
 	\item [$ \bullet $] EM \cite{SPML_2}: A SPML method, which maximizes the entropy of predicted probabilities for unknown labels. 
 	\item [$ \bullet $] EM$\_$APL \cite{SPML_2} An improved version of EM, introducing asymmetric tolerance in assigning positive and negative pseudo-labels.
 	\item [$ \bullet $]	SCOB \cite{SPML_5}: A SPML method recovering cross-object relationships by incorporating class activation as semantic guidance
 	\item [$ \bullet $] VLPL \cite{VLPL}: A SPML method, which is based on CLIP model to extract positive and pseudo-labels.
 	\item [$ \bullet $] PLCSL \cite{PLML_2}: A PML method, which estimate the class distribution through a dedicated temporary model.
 	\item [$ \bullet $] CLML \cite{CMLL}: An unbiased CML method, focusing on learning a multi-labeled classifier from complementary labeled data.
 \end{enumerate}
 To fairly compare with the state-of-the-art methods, all the comparative experiments were conducted in the same setting. Specifically, all methods use the same determined-labeled training data, the same networks and optimization strategy.
\subsubsection{Implementation details}
 All experiments are implemented based on PyTorch and run on a NVIDIA GeForce RTX 4090 GPU. To make fair comparisons with all comparative methods, we adopt the same image Swin-L-based RAM and ViT-B/16-based CLIP initialized by their published pretrained weights for image and text extraction. Due to limited training resources, we set the batch size to 128 for the VOC and CUB datasets, with image input resized to 384$\times$384. For the COCO and NUS datasets, the batch size is set to 256, and images are resized to 224$\times$224. Besides, we set total epochs as 80, and employ the AdamW optimizer with an initial learning rate of $1e-2$, and a weight decay parameter set to $5e-2$. In our experiments, we employ four widely-used evaluation criteria, including \textit{MAP}, \textit{ranking loss}, \textit{one error}, and \textit{coverage}. Further information about these evaluation metrics is available in \cite{evaluation, SPML_1}. In our experiments, we utilize thelimited supervision generated by the determined label to correct theoutput of the RAM model. Specifically, for the training instance associated with determined label $y^{\lambda}=1$, we set $p(y^{\lambda} = 1 \mid x) = 1$; for the training instance associated with determined label $y^{\lambda}=0$, we set $p(y^{\lambda} = 1 \mid x) = 0$.
\subsection{Comparisons Result}
 Table \ref{table_2} and Table \ref{table_3} reports the comparison results between the proposed method and existing state-of-the-art weakly multi-label learning approaches from various metrics widely used in multi-label learning. To fairly validate the effectiveness of the proposed method, we utilize their published codes of these comparative methods, and re-implement them with the same Swin-L-based RAM and ViT-B/16-based CLIP as ours, ensuring that the only variable is the algorithm. We also compare our method with partial multi-label learning (PML) methods (i.e., PLCSL \cite{PLML_2}) and complementary multi-label learning (CML) methods (i.e., CMLL \cite{CMLL}) using their published codes.
 
 As shown in Table \ref{table_2} and Table \ref{table_3}, the proposed method achieves the best performance over existing state-of-the-art approaches on all benchmark datasets in terms of \textit{MAP}, \textit{ranking loss}, \textit{one error}, and \textit{coverage}. From the results, it can be observed that: 1) Compared to traditional SPML methods based on loss functions, our approach demonstrates significant advantages across all experiments. This underscores the limitation of traditional SPML methods in effectively leveraging the prior knowledge of large-scale pre-trained vision-language models. 2) Compared to the state-of-the-art SPML methods based on large-scale pre-trained vision-language models, our approach exhibits comparable performance improvements. This highlights the potential for advancements in methods designed specifically for the CLIP model in multi-label tasks. For instance, the utilization of a pre-trained large-scale multi-label model, such as RAM, underscores the superiority of our approach. 3) Compared to existing state-of-the-art partial multi-label learning methods and complemenatry multi-label learning methods, our approach significantly outperforms the comparing methods across all metrics. Specifically, on the \textit{MAP} metric, our method achieves a maximal performance improvement of 5.0$\%$ (VOC), 5.8$\%$ (COCO), 5.1$\%$ (NUS), and 5.9$\%$ (CUB). Furthermore, our method consistently demonstrates similar performance improvements across other metrics.
\subsection{Ablation Study}

  \begin{table*}[!htbp]
	\centering
	\renewcommand{\arraystretch}{1.2}
	\caption{The effectiveness analysis of proposed risk-consistent (RC) loss and similarity-based prompt (SBP)}
	\begin{tabular}{l|c|c|c|c}%c表示文本居中，c的个数是列数
		\toprule[1pt]
		Method & MAP$\uparrow$ & One error$\downarrow$  & Ranking loss$\downarrow$ & Coverage$\downarrow$  \\
		\midrule
		RAM zero-shot & 0.909 & 0.063 & 0.012 & 0.047 \\
		\midrule
		RAM + MSE & 0.912$\pm$0.001 & 0.058$\pm$0.003 & 0.012$\pm$0.001 & 0.042$\pm$0.001 \\
		\midrule
		RAM + BCE & 0.910$\pm$0.002 & 0.060$\pm$0.001 & 0.012$\pm$0.000 & 0.042$\pm$0.001 \\
		\midrule
		RAM + RC & 0.916$\pm$0.000 & 0.047$\pm$0.000 & 0.011$\pm$0.001 & 0.041$\pm$0.001 \\
		\midrule
		RAM + RC + SBP & \textbf{0.921$\pm$0.001} & \textbf{0.046$\pm$0.001} & \textbf{0.011$\pm$0.000} & \textbf{0.041$\pm$0.001} \\
		\bottomrule[1pt]
	\end{tabular}
	\label{table_4}
\end{table*}

\begin{figure*}[!htbp]	
	\begin{adjustbox}{center}
		\subfigure[MAP $\uparrow$]{
			\begin{minipage}[b]{0.25\linewidth}
				\centering
				\includegraphics[width=1.85in]{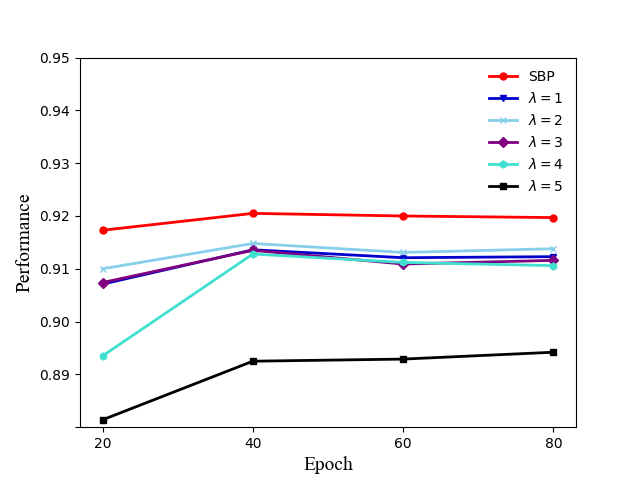}
				%\caption*{}
			\end{minipage}
		}%
		\subfigure[One error $\downarrow$]{
			\begin{minipage}[b]{0.25\linewidth}
				\centering
				\includegraphics[width=1.85in]{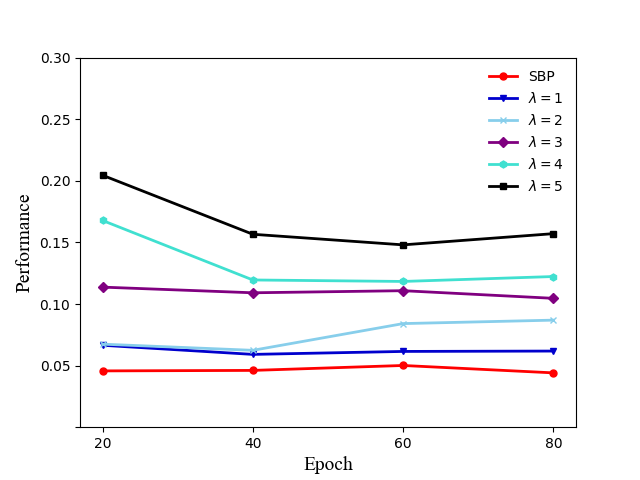}
				%\caption*{}
			\end{minipage}
		}%
		\subfigure[Ranking loss $\downarrow$]{
			\begin{minipage}[b]{0.25\linewidth}
				\centering
				\includegraphics[width=1.85in]{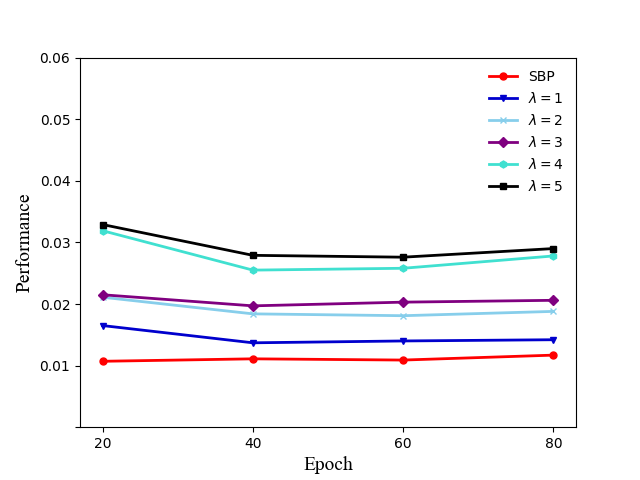}
				%\caption*{}
			\end{minipage}
		}%
		
		\subfigure[Coverage $\downarrow$]{
			\begin{minipage}[b]{0.25\linewidth}
				\centering
				\includegraphics[width=1.85in]{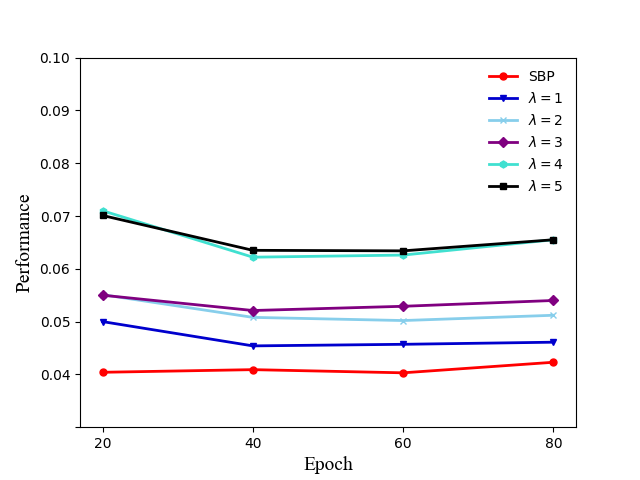}
				%\caption*{}
			\end{minipage}
		}%
	\end{adjustbox}
	
	\caption{Comparison results with \textit{RAM tag list} in terms of \textit{MAP} (the greater, the better), \textit{one error} (the smaller, the better),  \textit{ranking loss} (the smaller, the better), and \textit{coverage} (the smaller, the better).}
	\label{figure_3}
	\vspace{1em}
\end{figure*}

\begin{figure*}[!htbp]	
	\begin{adjustbox}{center}
		\subfigure[MAP $\uparrow$]{
			\begin{minipage}[b]{0.25\linewidth}
				\centering
				\includegraphics[width=1.85in]{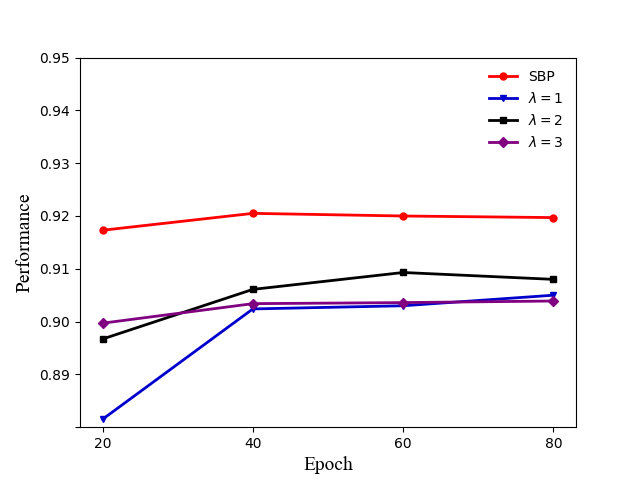}
				%\caption*{}
			\end{minipage}
		}%
		\subfigure[One error $\downarrow$]{
			\begin{minipage}[b]{0.25\linewidth}
				\centering
				\includegraphics[width=1.85in]{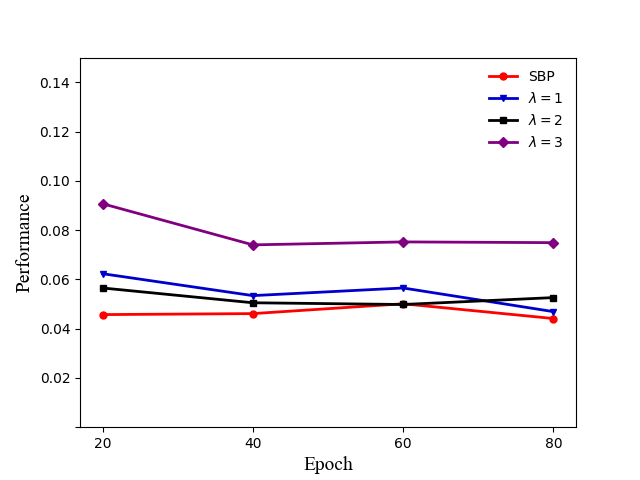}
				%\caption*{}
			\end{minipage}
		}%
		\subfigure[Ranking loss $\downarrow$]{
			\begin{minipage}[b]{0.25\linewidth}
				\centering
				\includegraphics[width=1.85in]{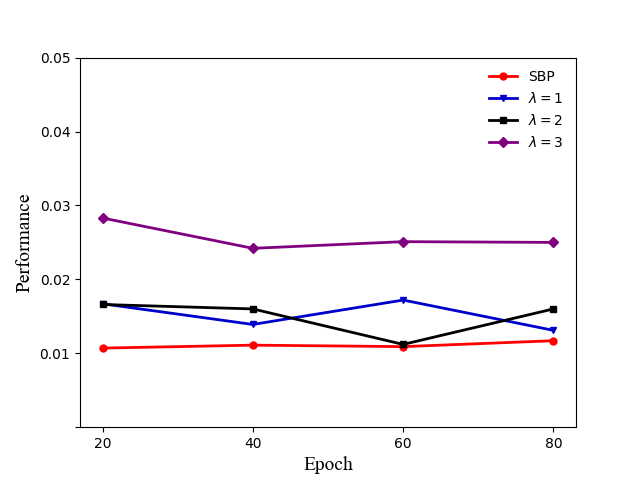}
				%\caption*{}
			\end{minipage}
		}%
		
		\subfigure[Coverage $\downarrow$]{
			\begin{minipage}[b]{0.25\linewidth}
				\centering
				\includegraphics[width=1.85in]{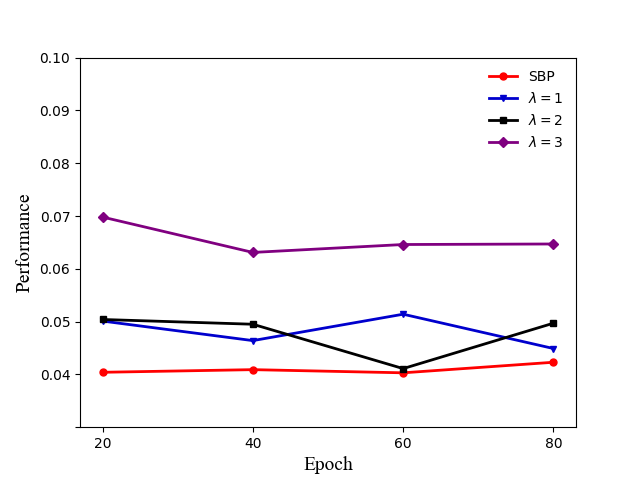}
				%\caption*{}
			\end{minipage}
		}%
	\end{adjustbox}
	
	\caption{Comparison results with \textit{open vocabulary} in terms of \textit{MAP} (the greater, the better), \textit{one error} (the smaller, the better),  \textit{ranking loss} (the smaller, the better), and \textit{coverage} (the smaller, the better).}
	\label{figure_4}
	\vspace{1em}
\end{figure*}
 \textbf{Effectiveness Analysis.} To validate the effectiveness of the  risk-consistent loss and similarity-based prompt proposed in the previous Section , we conduct additional ablation experiments on the VOC dataset. Specifically, we decomposed the proposed method into two parts, namely the risk-consistent loss (RC) and the supplemental similarity-based prompt learning (SBP) method. Subsequently, we incrementally incorporated these two parts into our method and compared them with RAM zero-shot and two common loss function, MSE and BCE. Experimental results are shown in Table \ref{table_4}. From Table \ref{table_4}, it can be observed that our proposed risk-consistent loss significantly improves over several baseline methods, and the introduction of similarity-based prompt further enhances the advantage of the proposed method. These results demonstrate the effectiveness of the two key components of our method.
 
 \textbf{Prompt Analysis based on RAM Tag List.} This paper introduces an supplemental similarity-based prompt. We adaptively calculate the proposed risk-consistent loss to obtain the optimal similar labels from RAM tag list for each label in the target label space $Y$.  To demonstrate the effectiveness of this part, we conducted additional ablation experiments with a fixed number of similar labels. The experimental results are shown in Figure \ref{figure_3}. Specifically, we select a fixed $\lambda$ number of similar labels from the RAM tag list for each candidate label, where $\lambda \leqslant \sigma$, and compare them with the proposed method. From Figure \ref{figure_3}, it can be seen that fixing the number of similar labels for each candidate label is not optimal. Too few similar labels do not significantly enhance semantic information, while too many introduce noise, leading to a decrease in performance. Fortunately, the proposed SBP can adaptively compute the number of similar labels, achieving the optimal performance.
 
 \textbf{Prompt Analysis based on Open Vocabulary.} The proposed SBP  utilize similar labels calculated based on the tag list provided by RAM, which has high generalization and is well-aligned with RAM's pre-training models. Additionally, we collected a list of similar labels from an open vocabulary. Specifically, we generated $\lambda$ similar labels for each candidate label using ChatGPT. These similar labels based solely on the output of the ChatGPT model without any constraints. Experimental results are shown in Figure \ref{figure_4}, indicating that the proposed SBP still achieves optimal performance. Unfortunately, this open-vovabulary-based approach did not yield desirable results, presumably due to mismatch with the RAM model, as we utilize the RAM image encoder to extract image features. If there are more open, large-scale, multi-label pre-trained models available, this open vocabulary approach might yield the expected results, which is also a problem worth further research.
\section{Conclusion}
 In this paper, we propose a novel labeling setting called Determined Multi-Label Learning (DMLL), which significantly reduces the labeling costs inherent in multi-label tasks. In this novel labeling setting, each training instance is assigned a \textit{determined label}, indicating whether the training instance contains the provided class label. The provided class label is randomly and uniformly chosen from the entire set of candidate labels. In this paper, we theoretically derive a risk-consistent estimator for learning a multi-label classifier from these determined-labeled training data. Additionally, we introduce a novel similarity-based prompt learning method, which minimizes the risk-consistent loss of large-scale pre-trained models to learn a supplemental prompt with richer semantic information. Extensive experimental resluts showcasing superior performance compared to existing state-of-the-art methods, demonstrating the efficacy of our method.

\bibliographystyle{ACM-Reference-Format}
\bibliography{sample-base}


\begin{thebibliography}{47}

%%% the Web address.

\ifx \showCODEN    \undefined \def \showCODEN     #1{\unskip}     \fi
\ifx \showDOI      \undefined \def \showDOI       #1{#1}\fi
\ifx \showISBNx    \undefined \def \showISBNx     #1{\unskip}     \fi
\ifx \showISBNxiii \undefined \def \showISBNxiii  #1{\unskip}     \fi
\ifx \showISSN     \undefined \def \showISSN      #1{\unskip}     \fi
\ifx \showLCCN     \undefined \def \showLCCN      #1{\unskip}     \fi
\ifx \shownote     \undefined \def \shownote      #1{#1}          \fi
\ifx \showarticletitle \undefined \def \showarticletitle #1{#1}   \fi
\ifx \showURL      \undefined \def \showURL       {\relax}        \fi
% invisible to TeX
\providecommand\bibfield[2]{#2}
\providecommand\bibinfo[2]{#2}
\providecommand\natexlab[1]{#1}
\providecommand\showeprint[2][]{arXiv:#2}

\bibitem[Ben-Baruch et~al\mbox{.}(2022)]%
        {PLML_2}
\bibfield{author}{\bibinfo{person}{Emanuel Ben-Baruch}, \bibinfo{person}{Tal
  Ridnik}, \bibinfo{person}{Itamar Friedman}, \bibinfo{person}{Avi Ben-Cohen},
  \bibinfo{person}{Nadav Zamir}, \bibinfo{person}{Asaf Noy}, {and}
  \bibinfo{person}{Lihi Zelnik-Manor}.} \bibinfo{year}{2022}\natexlab{}.
\newblock \showarticletitle{Multi-label classification with partial annotations
  using class-aware selective loss}. In \bibinfo{booktitle}{\emph{Proceedings
  of the IEEE/CVF Conference on Computer Vision and Pattern Recognition}}.
  \bibinfo{pages}{4764--4772}.
\newblock


\bibitem[Bucak et~al\mbox{.}(2011)]%
        {ML_14}
\bibfield{author}{\bibinfo{person}{Serhat~Selcuk Bucak}, \bibinfo{person}{Rong
  Jin}, {and} \bibinfo{person}{Anil~K Jain}.} \bibinfo{year}{2011}\natexlab{}.
\newblock \showarticletitle{Multi-label learning with incomplete class
  assignments}. In \bibinfo{booktitle}{\emph{CVPR 2011}}. IEEE,
  \bibinfo{pages}{2801--2808}.
\newblock


\bibitem[Chen et~al\mbox{.}(2023)]%
        {SPML_5}
\bibfield{author}{\bibinfo{person}{Cheng Chen}, \bibinfo{person}{Yifan Zhao},
  {and} \bibinfo{person}{Jia Li}.} \bibinfo{year}{2023}\natexlab{}.
\newblock \showarticletitle{Semantic Contrastive Bootstrapping for
  Single-Positive Multi-label Recognition}.
\newblock \bibinfo{journal}{\emph{International Journal of Computer Vision}}
  \bibinfo{volume}{131}, \bibinfo{number}{12} (\bibinfo{year}{2023}),
  \bibinfo{pages}{3289--3306}.
\newblock


\bibitem[Chen et~al\mbox{.}(2016)]%
        {ML_6}
\bibfield{author}{\bibinfo{person}{Di Chen}, \bibinfo{person}{Yexiang Xue},
  \bibinfo{person}{Shuo Chen}, \bibinfo{person}{Daniel Fink}, {and}
  \bibinfo{person}{Carla Gomes}.} \bibinfo{year}{2016}\natexlab{}.
\newblock \showarticletitle{Deep multi-species embedding}.
\newblock \bibinfo{journal}{\emph{arXiv preprint arXiv:1609.09353}}
  (\bibinfo{year}{2016}).
\newblock


\bibitem[Chua et~al\mbox{.}(2009)]%
        {NUS}
\bibfield{author}{\bibinfo{person}{Tat-Seng Chua}, \bibinfo{person}{Jinhui
  Tang}, \bibinfo{person}{Richang Hong}, \bibinfo{person}{Haojie Li},
  \bibinfo{person}{Zhiping Luo}, {and} \bibinfo{person}{Yantao Zheng}.}
  \bibinfo{year}{2009}\natexlab{}.
\newblock \showarticletitle{Nus-wide: a real-world web image database from
  national university of singapore}. In \bibinfo{booktitle}{\emph{Proceedings
  of the ACM international conference on image and video retrieval}}.
  \bibinfo{pages}{1--9}.
\newblock


\bibitem[Cole et~al\mbox{.}(2021)]%
        {SPML_1}
\bibfield{author}{\bibinfo{person}{Elijah Cole}, \bibinfo{person}{Oisin
  Mac~Aodha}, \bibinfo{person}{Titouan Lorieul}, \bibinfo{person}{Pietro
  Perona}, \bibinfo{person}{Dan Morris}, {and} \bibinfo{person}{Nebojsa
  Jojic}.} \bibinfo{year}{2021}\natexlab{}.
\newblock \showarticletitle{Multi-label learning from single positive labels}.
  In \bibinfo{booktitle}{\emph{Proceedings of the IEEE/CVF Conference on
  Computer Vision and Pattern Recognition}}. \bibinfo{pages}{933--942}.
\newblock


\bibitem[Deng et~al\mbox{.}(2014)]%
        {ML_7}
\bibfield{author}{\bibinfo{person}{Jia Deng}, \bibinfo{person}{Olga
  Russakovsky}, \bibinfo{person}{Jonathan Krause}, \bibinfo{person}{Michael~S
  Bernstein}, \bibinfo{person}{Alex Berg}, {and} \bibinfo{person}{Li Fei-Fei}.}
  \bibinfo{year}{2014}\natexlab{}.
\newblock \showarticletitle{Scalable multi-label annotation}. In
  \bibinfo{booktitle}{\emph{Proceedings of the SIGCHI Conference on Human
  Factors in Computing Systems}}. \bibinfo{pages}{3099--3102}.
\newblock


\bibitem[Ding et~al\mbox{.}(2023)]%
        {SCPnet}
\bibfield{author}{\bibinfo{person}{Zixuan Ding}, \bibinfo{person}{Ao Wang},
  \bibinfo{person}{Hui Chen}, \bibinfo{person}{Qiang Zhang},
  \bibinfo{person}{Pengzhang Liu}, \bibinfo{person}{Yongjun Bao},
  \bibinfo{person}{Weipeng Yan}, {and} \bibinfo{person}{Jungong Han}.}
  \bibinfo{year}{2023}\natexlab{}.
\newblock \showarticletitle{Exploring Structured Semantic Prior for Multi Label
  Recognition with Incomplete Labels}. In \bibinfo{booktitle}{\emph{Proceedings
  of the IEEE/CVF Conference on Computer Vision and Pattern Recognition}}.
  \bibinfo{pages}{3398--3407}.
\newblock


\bibitem[Everingham et~al\mbox{.}(2012)]%
        {VOC}
\bibfield{author}{\bibinfo{person}{M Everingham}, \bibinfo{person}{L Van~Gool},
  \bibinfo{person}{CKI Williams}, \bibinfo{person}{J Winn}, {and}
  \bibinfo{person}{A Zisserman}.} \bibinfo{year}{2012}\natexlab{}.
\newblock \showarticletitle{The PASCAL visual object classes challenge 2012
  (VOC2012) results. 2012 http://www. pascal-network. org/challenges}. In
  \bibinfo{booktitle}{\emph{VOC/voc2012/workshop/index. html}}.
\newblock


\bibitem[Feng et~al\mbox{.}(2020)]%
        {PLL1}
\bibfield{author}{\bibinfo{person}{Lei Feng}, \bibinfo{person}{Jiaqi Lv},
  \bibinfo{person}{Bo Han}, \bibinfo{person}{Miao Xu}, \bibinfo{person}{Gang
  Niu}, \bibinfo{person}{Xin Geng}, \bibinfo{person}{Bo An}, {and}
  \bibinfo{person}{Masashi Sugiyama}.} \bibinfo{year}{2020}\natexlab{}.
\newblock \showarticletitle{Provably consistent partial-label learning}.
\newblock \bibinfo{journal}{\emph{Advances in neural information processing
  systems}}  \bibinfo{volume}{33} (\bibinfo{year}{2020}),
  \bibinfo{pages}{10948--10960}.
\newblock


\bibitem[Gao et~al\mbox{.}(2020)]%
        {VLM6}
\bibfield{author}{\bibinfo{person}{Tianyu Gao}, \bibinfo{person}{Adam Fisch},
  {and} \bibinfo{person}{Danqi Chen}.} \bibinfo{year}{2020}\natexlab{}.
\newblock \showarticletitle{Making pre-trained language models better few-shot
  learners}.
\newblock \bibinfo{journal}{\emph{arXiv preprint arXiv:2012.15723}}
  (\bibinfo{year}{2020}).
\newblock


\bibitem[Gao et~al\mbox{.}(2023)]%
        {CMLL}
\bibfield{author}{\bibinfo{person}{Yi Gao}, \bibinfo{person}{Miao Xu}, {and}
  \bibinfo{person}{Min-Ling Zhang}.} \bibinfo{year}{2023}\natexlab{}.
\newblock \showarticletitle{Unbiased risk estimator to multi-labeled
  complementary label learning}. In \bibinfo{booktitle}{\emph{Proceedings of
  the Thirty-Second International Joint Conference on Artificial Intelligence,
  IJCAI}}. \bibinfo{pages}{3732--3740}.
\newblock


\bibitem[Gong et~al\mbox{.}(2013)]%
        {ML_10}
\bibfield{author}{\bibinfo{person}{Yunchao Gong}, \bibinfo{person}{Yangqing
  Jia}, \bibinfo{person}{Thomas Leung}, \bibinfo{person}{Alexander Toshev},
  {and} \bibinfo{person}{Sergey Ioffe}.} \bibinfo{year}{2013}\natexlab{}.
\newblock \showarticletitle{Deep convolutional ranking for multilabel image
  annotation}.
\newblock \bibinfo{journal}{\emph{arXiv preprint arXiv:1312.4894}}
  (\bibinfo{year}{2013}).
\newblock


\bibitem[He et~al\mbox{.}(2023)]%
        {MKT}
\bibfield{author}{\bibinfo{person}{Sunan He}, \bibinfo{person}{Taian Guo},
  \bibinfo{person}{Tao Dai}, \bibinfo{person}{Ruizhi Qiao},
  \bibinfo{person}{Xiujun Shu}, \bibinfo{person}{Bo Ren}, {and}
  \bibinfo{person}{Shu-Tao Xia}.} \bibinfo{year}{2023}\natexlab{}.
\newblock \showarticletitle{Open-vocabulary multi-label classification via
  multi-modal knowledge transfer}. In \bibinfo{booktitle}{\emph{Proceedings of
  the AAAI Conference on Artificial Intelligence}}, Vol.~\bibinfo{volume}{37}.
  \bibinfo{pages}{808--816}.
\newblock


\bibitem[Hsu et~al\mbox{.}(2009)]%
        {ML_9}
\bibfield{author}{\bibinfo{person}{Daniel~J Hsu}, \bibinfo{person}{Sham~M
  Kakade}, \bibinfo{person}{John Langford}, {and} \bibinfo{person}{Tong
  Zhang}.} \bibinfo{year}{2009}\natexlab{}.
\newblock \showarticletitle{Multi-label prediction via compressed sensing}.
\newblock \bibinfo{journal}{\emph{Advances in neural information processing
  systems}}  \bibinfo{volume}{22} (\bibinfo{year}{2009}).
\newblock


\bibitem[Jia et~al\mbox{.}(2021)]%
        {VLM1}
\bibfield{author}{\bibinfo{person}{Chao Jia}, \bibinfo{person}{Yinfei Yang},
  \bibinfo{person}{Ye Xia}, \bibinfo{person}{Yi-Ting Chen},
  \bibinfo{person}{Zarana Parekh}, \bibinfo{person}{Hieu Pham},
  \bibinfo{person}{Quoc Le}, \bibinfo{person}{Yun-Hsuan Sung},
  \bibinfo{person}{Zhen Li}, {and} \bibinfo{person}{Tom Duerig}.}
  \bibinfo{year}{2021}\natexlab{}.
\newblock \showarticletitle{Scaling up visual and vision-language
  representation learning with noisy text supervision}. In
  \bibinfo{booktitle}{\emph{International conference on machine learning}}.
  PMLR, \bibinfo{pages}{4904--4916}.
\newblock


\bibitem[Jiang et~al\mbox{.}(2020)]%
        {VLM7}
\bibfield{author}{\bibinfo{person}{Zhengbao Jiang}, \bibinfo{person}{Frank~F
  Xu}, \bibinfo{person}{Jun Araki}, {and} \bibinfo{person}{Graham Neubig}.}
  \bibinfo{year}{2020}\natexlab{}.
\newblock \showarticletitle{How can we know what language models know?}
\newblock \bibinfo{journal}{\emph{Transactions of the Association for
  Computational Linguistics}}  \bibinfo{volume}{8} (\bibinfo{year}{2020}),
  \bibinfo{pages}{423--438}.
\newblock


\bibitem[Kim et~al\mbox{.}(2022)]%
        {SPML_3}
\bibfield{author}{\bibinfo{person}{Youngwook Kim}, \bibinfo{person}{Jae~Myung
  Kim}, \bibinfo{person}{Zeynep Akata}, {and} \bibinfo{person}{Jungwoo Lee}.}
  \bibinfo{year}{2022}\natexlab{}.
\newblock \showarticletitle{Large loss matters in weakly supervised multi-label
  classification}. In \bibinfo{booktitle}{\emph{Proceedings of the IEEE/CVF
  Conference on Computer Vision and Pattern Recognition}}.
  \bibinfo{pages}{14156--14165}.
\newblock


\bibitem[Li et~al\mbox{.}(2023)]%
        {VLM2}
\bibfield{author}{\bibinfo{person}{Junnan Li}, \bibinfo{person}{Dongxu Li},
  \bibinfo{person}{Silvio Savarese}, {and} \bibinfo{person}{Steven Hoi}.}
  \bibinfo{year}{2023}\natexlab{}.
\newblock \showarticletitle{Blip-2: Bootstrapping language-image pre-training
  with frozen image encoders and large language models}.
\newblock \bibinfo{journal}{\emph{arXiv preprint arXiv:2301.12597}}
  (\bibinfo{year}{2023}).
\newblock


\bibitem[Lin et~al\mbox{.}(2014)]%
        {COCO}
\bibfield{author}{\bibinfo{person}{Tsung-Yi Lin}, \bibinfo{person}{Michael
  Maire}, \bibinfo{person}{Serge Belongie}, \bibinfo{person}{James Hays},
  \bibinfo{person}{Pietro Perona}, \bibinfo{person}{Deva Ramanan},
  \bibinfo{person}{Piotr Doll{\'a}r}, {and} \bibinfo{person}{C~Lawrence
  Zitnick}.} \bibinfo{year}{2014}\natexlab{}.
\newblock \showarticletitle{Microsoft coco: Common objects in context}. In
  \bibinfo{booktitle}{\emph{Computer Vision--ECCV 2014: 13th European
  Conference, Zurich, Switzerland, September 6-12, 2014, Proceedings, Part V
  13}}. Springer, \bibinfo{pages}{740--755}.
\newblock


\bibitem[Ma et~al\mbox{.}(2022)]%
        {VLM8}
\bibfield{author}{\bibinfo{person}{Chaofan Ma}, \bibinfo{person}{Yuhuan Yang},
  \bibinfo{person}{Yanfeng Wang}, \bibinfo{person}{Ya Zhang}, {and}
  \bibinfo{person}{Weidi Xie}.} \bibinfo{year}{2022}\natexlab{}.
\newblock \showarticletitle{Open-vocabulary semantic segmentation with frozen
  vision-language models}.
\newblock \bibinfo{journal}{\emph{arXiv preprint arXiv:2210.15138}}
  (\bibinfo{year}{2022}).
\newblock


\bibitem[Mac~Aodha et~al\mbox{.}(2019)]%
        {ML_15}
\bibfield{author}{\bibinfo{person}{Oisin Mac~Aodha}, \bibinfo{person}{Elijah
  Cole}, {and} \bibinfo{person}{Pietro Perona}.}
  \bibinfo{year}{2019}\natexlab{}.
\newblock \showarticletitle{Presence-only geographical priors for fine-grained
  image classification}. In \bibinfo{booktitle}{\emph{Proceedings of the
  IEEE/CVF International Conference on Computer Vision}}.
  \bibinfo{pages}{9596--9606}.
\newblock


\bibitem[Radford et~al\mbox{.}(2021a)]%
        {VLM3}
\bibfield{author}{\bibinfo{person}{Alec Radford}, \bibinfo{person}{Jong~Wook
  Kim}, \bibinfo{person}{Chris Hallacy}, \bibinfo{person}{Aditya Ramesh},
  \bibinfo{person}{Gabriel Goh}, \bibinfo{person}{Sandhini Agarwal},
  \bibinfo{person}{Girish Sastry}, \bibinfo{person}{Amanda Askell},
  \bibinfo{person}{Pamela Mishkin}, \bibinfo{person}{Jack Clark},
  {et~al\mbox{.}}} \bibinfo{year}{2021}\natexlab{a}.
\newblock \showarticletitle{Learning transferable visual models from natural
  language supervision}. In \bibinfo{booktitle}{\emph{International conference
  on machine learning}}. PMLR, \bibinfo{pages}{8748--8763}.
\newblock


\bibitem[Radford et~al\mbox{.}(2021b)]%
        {CLIP}
\bibfield{author}{\bibinfo{person}{Alec Radford}, \bibinfo{person}{Jong~Wook
  Kim}, \bibinfo{person}{Chris Hallacy}, \bibinfo{person}{Aditya Ramesh},
  \bibinfo{person}{Gabriel Goh}, \bibinfo{person}{Sandhini Agarwal},
  \bibinfo{person}{Girish Sastry}, \bibinfo{person}{Amanda Askell},
  \bibinfo{person}{Pamela Mishkin}, \bibinfo{person}{Jack Clark},
  {et~al\mbox{.}}} \bibinfo{year}{2021}\natexlab{b}.
\newblock \showarticletitle{Learning transferable visual models from natural
  language supervision}. In \bibinfo{booktitle}{\emph{International conference
  on machine learning}}. PMLR, \bibinfo{pages}{8748--8763}.
\newblock


\bibitem[Shin et~al\mbox{.}(2020)]%
        {VLM9}
\bibfield{author}{\bibinfo{person}{Taylor Shin}, \bibinfo{person}{Yasaman
  Razeghi}, \bibinfo{person}{Robert~L Logan~IV}, \bibinfo{person}{Eric
  Wallace}, {and} \bibinfo{person}{Sameer Singh}.}
  \bibinfo{year}{2020}\natexlab{}.
\newblock \showarticletitle{Autoprompt: Eliciting knowledge from language
  models with automatically generated prompts}.
\newblock \bibinfo{journal}{\emph{arXiv preprint arXiv:2010.15980}}
  (\bibinfo{year}{2020}).
\newblock


\bibitem[Sun et~al\mbox{.}(2022)]%
        {DUALCoOp}
\bibfield{author}{\bibinfo{person}{Ximeng Sun}, \bibinfo{person}{Ping Hu},
  {and} \bibinfo{person}{Kate Saenko}.} \bibinfo{year}{2022}\natexlab{}.
\newblock \showarticletitle{Dualcoop: Fast adaptation to multi-label
  recognition with limited annotations}.
\newblock \bibinfo{journal}{\emph{Advances in Neural Information Processing
  Systems}}  \bibinfo{volume}{35} (\bibinfo{year}{2022}),
  \bibinfo{pages}{30569--30582}.
\newblock


\bibitem[Sun et~al\mbox{.}(2010)]%
        {ML_13}
\bibfield{author}{\bibinfo{person}{Yu-Yin Sun}, \bibinfo{person}{Yin Zhang},
  {and} \bibinfo{person}{Zhi-Hua Zhou}.} \bibinfo{year}{2010}\natexlab{}.
\newblock \showarticletitle{Multi-label learning with weak label}. In
  \bibinfo{booktitle}{\emph{Proceedings of the AAAI conference on artificial
  intelligence}}, Vol.~\bibinfo{volume}{24}. \bibinfo{pages}{593--598}.
\newblock


\bibitem[Tang et~al\mbox{.}(2020)]%
        {ML_4}
\bibfield{author}{\bibinfo{person}{Pingjie Tang}, \bibinfo{person}{Meng Jiang},
  \bibinfo{person}{Bryan~Ning Xia}, \bibinfo{person}{Jed~W Pitera},
  \bibinfo{person}{Jeffrey Welser}, {and} \bibinfo{person}{Nitesh~V Chawla}.}
  \bibinfo{year}{2020}\natexlab{}.
\newblock \showarticletitle{Multi-label patent categorization with non-local
  attention-based graph convolutional network}. In
  \bibinfo{booktitle}{\emph{Proceedings of the AAAI Conference on Artificial
  Intelligence}}, Vol.~\bibinfo{volume}{34}. \bibinfo{pages}{9024--9031}.
\newblock


\bibitem[Vallet and Sakamoto(2015)]%
        {ML_3}
\bibfield{author}{\bibinfo{person}{Alexis Vallet} {and}
  \bibinfo{person}{Hiroyasu Sakamoto}.} \bibinfo{year}{2015}\natexlab{}.
\newblock \showarticletitle{A multi-label convolutional neural network for
  automatic image annotation}.
\newblock \bibinfo{journal}{\emph{Journal of information processing}}
  \bibinfo{volume}{23}, \bibinfo{number}{6} (\bibinfo{year}{2015}),
  \bibinfo{pages}{767--775}.
\newblock


\bibitem[Wah et~al\mbox{.}(2011)]%
        {CUB}
\bibfield{author}{\bibinfo{person}{Catherine Wah}, \bibinfo{person}{Steve
  Branson}, \bibinfo{person}{Peter Welinder}, \bibinfo{person}{Pietro Perona},
  {and} \bibinfo{person}{Serge Belongie}.} \bibinfo{year}{2011}\natexlab{}.
\newblock \showarticletitle{The caltech-ucsd birds-200-2011 dataset}.
\newblock  (\bibinfo{year}{2011}).
\newblock


\bibitem[Wang et~al\mbox{.}(2023)]%
        {HSPNet}
\bibfield{author}{\bibinfo{person}{Ao Wang}, \bibinfo{person}{Hui Chen},
  \bibinfo{person}{Zijia Lin}, \bibinfo{person}{Zixuan Ding},
  \bibinfo{person}{Pengzhang Liu}, \bibinfo{person}{Yongjun Bao},
  \bibinfo{person}{Weipeng Yan}, {and} \bibinfo{person}{Guiguang Ding}.}
  \bibinfo{year}{2023}\natexlab{}.
\newblock \showarticletitle{Hierarchical Prompt Learning Using CLIP for
  Multi-label Classification with Single Positive Labels}. In
  \bibinfo{booktitle}{\emph{Proceedings of the 31st ACM International
  Conference on Multimedia}}. \bibinfo{pages}{5594--5604}.
\newblock


\bibitem[Wang et~al\mbox{.}(2016)]%
        {ML_12}
\bibfield{author}{\bibinfo{person}{Jiang Wang}, \bibinfo{person}{Yi Yang},
  \bibinfo{person}{Junhua Mao}, \bibinfo{person}{Zhiheng Huang},
  \bibinfo{person}{Chang Huang}, {and} \bibinfo{person}{Wei Xu}.}
  \bibinfo{year}{2016}\natexlab{}.
\newblock \showarticletitle{Cnn-rnn: A unified framework for multi-label image
  classification}. In \bibinfo{booktitle}{\emph{Proceedings of the IEEE
  conference on computer vision and pattern recognition}}.
  \bibinfo{pages}{2285--2294}.
\newblock


\bibitem[Wang et~al\mbox{.}(2021)]%
        {VLM4}
\bibfield{author}{\bibinfo{person}{Zirui Wang}, \bibinfo{person}{Jiahui Yu},
  \bibinfo{person}{Adams~Wei Yu}, \bibinfo{person}{Zihang Dai},
  \bibinfo{person}{Yulia Tsvetkov}, {and} \bibinfo{person}{Yuan Cao}.}
  \bibinfo{year}{2021}\natexlab{}.
\newblock \showarticletitle{Simvlm: Simple visual language model pretraining
  with weak supervision}.
\newblock \bibinfo{journal}{\emph{arXiv preprint arXiv:2108.10904}}
  (\bibinfo{year}{2021}).
\newblock


\bibitem[Wei et~al\mbox{.}(2015)]%
        {ML_11}
\bibfield{author}{\bibinfo{person}{Yunchao Wei}, \bibinfo{person}{Wei Xia},
  \bibinfo{person}{Min Lin}, \bibinfo{person}{Junshi Huang},
  \bibinfo{person}{Bingbing Ni}, \bibinfo{person}{Jian Dong},
  \bibinfo{person}{Yao Zhao}, {and} \bibinfo{person}{Shuicheng Yan}.}
  \bibinfo{year}{2015}\natexlab{}.
\newblock \showarticletitle{HCP: A flexible CNN framework for multi-label image
  classification}.
\newblock \bibinfo{journal}{\emph{IEEE transactions on pattern analysis and
  machine intelligence}} \bibinfo{volume}{38}, \bibinfo{number}{9}
  (\bibinfo{year}{2015}), \bibinfo{pages}{1901--1907}.
\newblock


\bibitem[Wu et~al\mbox{.}(2014)]%
        {ML_5}
\bibfield{author}{\bibinfo{person}{Bin Wu}, \bibinfo{person}{Erheng Zhong},
  \bibinfo{person}{Andrew Horner}, {and} \bibinfo{person}{Qiang Yang}.}
  \bibinfo{year}{2014}\natexlab{}.
\newblock \showarticletitle{Music emotion recognition by multi-label
  multi-layer multi-instance multi-view learning}. In
  \bibinfo{booktitle}{\emph{Proceedings of the 22nd ACM international
  conference on Multimedia}}. \bibinfo{pages}{117--126}.
\newblock


\bibitem[Xie et~al\mbox{.}(2021)]%
        {PLML_1}
\bibfield{author}{\bibinfo{person}{Ming-Kun Xie}, \bibinfo{person}{Feng Sun},
  {and} \bibinfo{person}{Sheng-Jun Huang}.} \bibinfo{year}{2021}\natexlab{}.
\newblock \showarticletitle{Partial multi-label learning with meta
  disambiguation}. In \bibinfo{booktitle}{\emph{Proceedings of the 27th ACM
  SIGKDD conference on knowledge discovery \& data mining}}.
  \bibinfo{pages}{1904--1912}.
\newblock


\bibitem[Xing et~al\mbox{.}(2023)]%
        {VLPL}
\bibfield{author}{\bibinfo{person}{Xin Xing}, \bibinfo{person}{Zhexiao Xiong},
  \bibinfo{person}{Abby Stylianou}, \bibinfo{person}{Srikumar Sastry},
  \bibinfo{person}{Liyu Gong}, {and} \bibinfo{person}{Nathan Jacobs}.}
  \bibinfo{year}{2023}\natexlab{}.
\newblock \showarticletitle{Vision-Language Pseudo-Labels for Single-Positive
  Multi-Label Learning}.
\newblock \bibinfo{journal}{\emph{arXiv preprint arXiv:2310.15985}}
  (\bibinfo{year}{2023}).
\newblock


\bibitem[Xu et~al\mbox{.}(2019)]%
        {ML_1}
\bibfield{author}{\bibinfo{person}{Ning Xu}, \bibinfo{person}{Yun-Peng Liu},
  {and} \bibinfo{person}{Xin Geng}.} \bibinfo{year}{2019}\natexlab{}.
\newblock \showarticletitle{Label enhancement for label distribution learning}.
\newblock \bibinfo{journal}{\emph{IEEE Transactions on Knowledge and Data
  Engineering}} \bibinfo{volume}{33}, \bibinfo{number}{4}
  (\bibinfo{year}{2019}), \bibinfo{pages}{1632--1643}.
\newblock


\bibitem[Xu et~al\mbox{.}(2022b)]%
        {SPML_4}
\bibfield{author}{\bibinfo{person}{Ning Xu}, \bibinfo{person}{Congyu Qiao},
  \bibinfo{person}{Jiaqi Lv}, \bibinfo{person}{Xin Geng}, {and}
  \bibinfo{person}{Min-Ling Zhang}.} \bibinfo{year}{2022}\natexlab{b}.
\newblock \showarticletitle{One positive label is sufficient: Single-positive
  multi-label learning with label enhancement}.
\newblock \bibinfo{journal}{\emph{Advances in Neural Information Processing
  Systems}}  \bibinfo{volume}{35} (\bibinfo{year}{2022}),
  \bibinfo{pages}{21765--21776}.
\newblock


\bibitem[Xu et~al\mbox{.}(2022a)]%
        {VLM10}
\bibfield{author}{\bibinfo{person}{Shichao Xu}, \bibinfo{person}{Yikang Li},
  \bibinfo{person}{Jenhao Hsiao}, \bibinfo{person}{Chiuman Ho}, {and}
  \bibinfo{person}{Zhu Qi}.} \bibinfo{year}{2022}\natexlab{a}.
\newblock \showarticletitle{A Dual Modality Approach For (Zero-Shot)
  Multi-Label Classification}.
\newblock \bibinfo{journal}{\emph{arXiv preprint arXiv:2208.09562}}
  (\bibinfo{year}{2022}).
\newblock


\bibitem[Yu et~al\mbox{.}(2014)]%
        {ML_2}
\bibfield{author}{\bibinfo{person}{Hsiang-Fu Yu}, \bibinfo{person}{Prateek
  Jain}, \bibinfo{person}{Purushottam Kar}, {and} \bibinfo{person}{Inderjit
  Dhillon}.} \bibinfo{year}{2014}\natexlab{}.
\newblock \showarticletitle{Large-scale multi-label learning with missing
  labels}. In \bibinfo{booktitle}{\emph{International conference on machine
  learning}}. PMLR, \bibinfo{pages}{593--601}.
\newblock


\bibitem[Yu et~al\mbox{.}(2022)]%
        {VLM5}
\bibfield{author}{\bibinfo{person}{Jiahui Yu}, \bibinfo{person}{Zirui Wang},
  \bibinfo{person}{Vijay Vasudevan}, \bibinfo{person}{Legg Yeung},
  \bibinfo{person}{Mojtaba Seyedhosseini}, {and} \bibinfo{person}{Yonghui Wu}.}
  \bibinfo{year}{2022}\natexlab{}.
\newblock \showarticletitle{Coca: Contrastive captioners are image-text
  foundation models}.
\newblock \bibinfo{journal}{\emph{arXiv preprint arXiv:2205.01917}}
  (\bibinfo{year}{2022}).
\newblock


\bibitem[Zhang and Zhou(2007)]%
        {ML_8}
\bibfield{author}{\bibinfo{person}{Min-Ling Zhang} {and}
  \bibinfo{person}{Zhi-Hua Zhou}.} \bibinfo{year}{2007}\natexlab{}.
\newblock \showarticletitle{ML-KNN: A lazy learning approach to multi-label
  learning}.
\newblock \bibinfo{journal}{\emph{Pattern recognition}} \bibinfo{volume}{40},
  \bibinfo{number}{7} (\bibinfo{year}{2007}), \bibinfo{pages}{2038--2048}.
\newblock


\bibitem[Zhang and Zhou(2013)]%
        {evaluation}
\bibfield{author}{\bibinfo{person}{Min-Ling Zhang} {and}
  \bibinfo{person}{Zhi-Hua Zhou}.} \bibinfo{year}{2013}\natexlab{}.
\newblock \showarticletitle{A review on multi-label learning algorithms}.
\newblock \bibinfo{journal}{\emph{IEEE transactions on knowledge and data
  engineering}} \bibinfo{volume}{26}, \bibinfo{number}{8}
  (\bibinfo{year}{2013}), \bibinfo{pages}{1819--1837}.
\newblock


\bibitem[Zhang et~al\mbox{.}(2023)]%
        {RAM}
\bibfield{author}{\bibinfo{person}{Youcai Zhang}, \bibinfo{person}{Xinyu
  Huang}, \bibinfo{person}{Jinyu Ma}, \bibinfo{person}{Zhaoyang Li},
  \bibinfo{person}{Zhaochuan Luo}, \bibinfo{person}{Yanchun Xie},
  \bibinfo{person}{Yuzhuo Qin}, \bibinfo{person}{Tong Luo},
  \bibinfo{person}{Yaqian Li}, \bibinfo{person}{Shilong Liu}, {et~al\mbox{.}}}
  \bibinfo{year}{2023}\natexlab{}.
\newblock \showarticletitle{Recognize Anything: A Strong Image Tagging Model}.
\newblock \bibinfo{journal}{\emph{arXiv preprint arXiv:2306.03514}}
  (\bibinfo{year}{2023}).
\newblock


\bibitem[Zhou et~al\mbox{.}(2022a)]%
        {SPML_2}
\bibfield{author}{\bibinfo{person}{Donghao Zhou}, \bibinfo{person}{Pengfei
  Chen}, \bibinfo{person}{Qiong Wang}, \bibinfo{person}{Guangyong Chen}, {and}
  \bibinfo{person}{Pheng-Ann Heng}.} \bibinfo{year}{2022}\natexlab{a}.
\newblock \showarticletitle{Acknowledging the unknown for multi-label learning
  with single positive labels}. In \bibinfo{booktitle}{\emph{European
  Conference on Computer Vision}}. Springer, \bibinfo{pages}{423--440}.
\newblock


\bibitem[Zhou et~al\mbox{.}(2022b)]%
        {CoCoOp}
\bibfield{author}{\bibinfo{person}{Kaiyang Zhou}, \bibinfo{person}{Jingkang
  Yang}, \bibinfo{person}{Chen~Change Loy}, {and} \bibinfo{person}{Ziwei Liu}.}
  \bibinfo{year}{2022}\natexlab{b}.
\newblock \showarticletitle{Conditional prompt learning for vision-language
  models}. In \bibinfo{booktitle}{\emph{Proceedings of the IEEE/CVF Conference
  on Computer Vision and Pattern Recognition}}. \bibinfo{pages}{16816--16825}.
\newblock


\end{thebibliography}

\appendix
\section{Appdendix}
\subsection{Details of Eq.(5)}
\begin{equation}\label{eq11}
	\begin{aligned}
		& \sum_{Y \in \mathcal{C}} \mathcal{L}(f(x), Y)p(Y\mid x)\\
		& = \sum_{Y \in \mathcal{C}} \sum_{j \in Y} \ell^jp(Y\mid x) +  \sum_{Y \in \mathcal{C}} \sum_{j \in Y} \bar{\ell}^jp(Y\mid x) \\
		& = \sum_{j=1}^{k} \ell^j \sum_{Y \in \mathcal{C}_j}p(Y\mid x) + \sum_{j=1}^{k}\bar{\ell}\sum{Y \notin \mathcal{C}_j} p(Y \mid x) \\
		& = \sum_{j=1}^{k}p(y^j=1 \mid x)\ell^j \sum_{Y \in \mathcal{C}_j} \left[ \prod_{z\in Y, z\neq j}p(y^z=1\mid x) \prod_{z \notin Y}\left(1-p(y^z=1 \mid x)\right) \right] \\
		& + \sum_{j=1}^{k}\left(1 - p(y^j=1 \mid x)\right)\bar{\ell}^j \sum_{Y \notin \mathcal{C}_j} \left[ \prod_{z\in Y}p(y^z=1\mid x) \prod_{z \notin Y, z \neq j}\left(1-p(y^z=1 \mid x)\right) \right] \\
		& = \sum_{j=1}^{k}\left[p(y^j=1 \mid x) \ell_j + \left(1-p(y^j=1 \mid x)\right) \bar{\ell}^j \right],
	\end{aligned}
	\nonumber
\end{equation}
where $\mathcal{C}_j$ denotes the subset of $\mathcal{C}$ which contains label $j$.
\end{document}